\pdfoutput=1

\documentclass[11pt]{article}
\usepackage{multirow}

\usepackage[final]{acl}
\usepackage[russian,english]{babel}
\usepackage{times}
\usepackage{latexsym}

\usepackage[T1]{fontenc}

\usepackage[utf8]{inputenc}

\usepackage{microtype}

\usepackage{inconsolata}

\usepackage{graphicx}

\title{Pisets: A Robust Speech Recognition System for Lectures and Interviews}

\author{
 \textbf{Ivan Bondarenko\textsuperscript{1}},
 \textbf{Daniil Grebenkin\textsuperscript{1,2}},
 \textbf{Oleg Sedukhin\textsuperscript{2}},
 \textbf{Mikhail Klementev\textsuperscript{1,2}},
\\
 \textbf{Roman Derunets\textsuperscript{1,2}},
 \textbf{Lyudmila Budneva\textsuperscript{1}}
\\
 \textsuperscript{1}Novosibirsk State University,
 \textsuperscript{2}Siberian Neuronets LLC
\\
 \small{
   \textbf{Correspondence:} \href{mailto:email@domain}{i.bondarenko@g.nsu.ru}
 }
}

\begin{document}
\maketitle
\begin{abstract}
This work presents a speech-to-text system "Pisets" for scientists and journalists which is based on a three-component architecture aimed at improving speech recognition accuracy while minimizing errors and hallucinations associated with the Whisper model. The architecture comprises primary recognition using Wav2Vec2, false positive filtering via the Audio Spectrogram Transformer (AST), and final speech recognition through Whisper. The implementation of curriculum learning methods and the utilization of diverse Russian-language speech corpora significantly enhanced the system's effectiveness. Additionally, advanced uncertainty modeling techniques were introduced, contributing to further improvements in transcription quality. The proposed approaches ensure robust transcribing of long audio data across various acoustic conditions compared to WhisperX and the usual Whisper model. The source code of "Pisets" system is publicly available at GitHub: \url{https://github.com/bond005/pisets}.
\end{abstract}

\section{Introduction}

Sustainable speech recognition systems are essential for scientists, journalists, and anyone processing audio recordings of interviews and meetings. They not only streamline transcription but also improve the reliability and accuracy of the output, facilitating better decision-making and communication.

We present the three-component architecture of the offline speech recognition system designed to enhance speech recognition accuracy while minimizing errors and hallucinations associated with the Whisper model. The architecture consists of three key components: primary recognition based on Wav2Vec2, false positive filtering using the Audio Spectrogram Transformer (AST), and final speech recognition utilizing Whisper. 

We called this system "Pisets" (in Russian, scribe), because it, like the ancient Roman scribe Tiro after Cicero, shorthand recordings of scientific speeches, interviews and other conversations.

\subsection{Primary Recognition Based on Wav2Vec2} 

The first component of our architecture relies on the Wav2Vec2 model \cite{Baevski2020wav2vec2A}, which effectively identifies the boundaries of the speech-containing segments. Unlike standard Voice Activity Detection (VAD) methods, which may be less sensitive and accurate, Wav2Vec2 offers a more powerful approach, which we refer to as VAD “on steroids”. This model has been trained on large volumes of audio data and leverages contextual information to more accurately determine the presence of speech segments.

To enhance Russian language recognition, we used a curriculum learning approach, which progressively increases task complexity during training. This method is informed by the “Formal Theory of Fun, Creativity, and Intrinsic Motivation.”~\cite{5508364}. In our context, complexity is characterized by the diversity of input audio data, including various accents, background noise, and acoustic conditions. We started with simpler, well-annotated data and gradually introduced more complex examples, which helped the model manage a wider range of speech fragments. Our model was trained using this curriculum learning strategy ~\cite{bengio2009curriculum} on open Russian-language speech corpora, including Golos \cite{Karpov2021GolosRD}, Russian Librispeech \cite{LibreSpeech}, RuDevices \cite{sova2021rudevices}, is publicly available at the Huggingface.

\subsection{False Positive Filtering Using the Audio Spectrogram Transformer (AST)} 

The second component of the architecture focuses on filtering false positive outputs generated by the speech detector. We selected the Audio Spectrogram Transformer (AST) \cite{gong2021ast}, trained on the Audioset ontology \cite{gemmeke2017audio}, due to its exceptional effectiveness in audio signal classification. Its implementation enables a reduction in the number of non-existent speech fragments that may be misinterpreted as actual speech. AST provides a deeper analysis of audio signals, highlighting significant acoustic features, which is particularly beneficial in noisy environments or complex acoustic spatial conditions. 

\subsection{Final Speech Recognition Using Whisper} 

The final component involves employing the Whisper model \cite{radford2023robust} to carry out the concluding stage of speech recognition. Whisper has demonstrated outstanding performance in various speech recognition tasks, and within our architecture, it plays the role of interpreting audio files that have undergone preliminary processing informed by the results of the first two components.

To enhance recognition accuracy in our system, we applied the BIRM (Bayesian Invariant Risk Minimization) algorithm \cite{Lin_2022_CVPR} and developed a speech environment concept. Constructing this environment involved creating an annotated speech corpora with a minimal error rate, allowing the Whisper model to better tackle the recognition task. Our training environment accounted for both the quality of annotations and the diversity of audio signals, resulting in a significant improvement in recognition outcomes. The resulting model is also available under the Apache 2.0 license on the Hugginface. We utilized three diverse speech corpora to enhance training across distinct linguistic and acoustic environments: Russian Librispeech \cite{LibreSpeech}, Taiga Speech~\cite{shavrina2017methodology}, Podlodka Speech \cite{podlodka_speech}.

\begin{figure}[t]
  \centering
  \includegraphics[width=0.7\columnwidth]{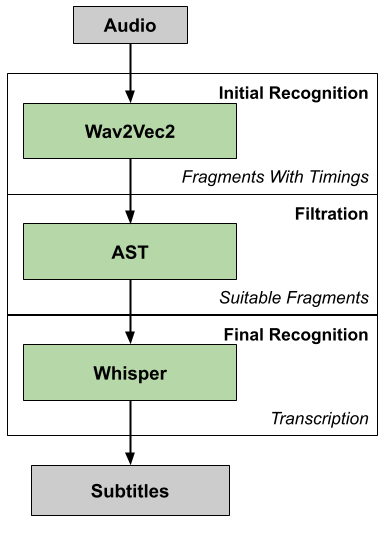}
  \caption{Proposed three-component speech recognition architecture}
  \label{fig:architecture}
\end{figure}

In conclusion, the proposed three-component architecture significantly reduces errors and hallucinations in speech recognition (see Fig. \ref{fig:architecture}). Each component plays a vital role in the overall process, creating a transformation chain from initial recognition to final output, ultimately leading to enhanced overall system effectiveness.

\section{Related Works}

The development of automated transcription systems for lectures and interviews relies critically on speech recognition methodologies. Beyond the fundamental task of acoustic-to-text conversion, such systems must address ancillary linguistic processing challenges to ensure output fidelity. These include punctuation restoration, capitalization recovery, numeral normalization, and syntactic disambiguation—operations essential for producing human-interpretable transcripts. Historically, these subtasks were addressed through modular subsystems: for instance, Kaldi-based frameworks employing classical Hidden Markov Model-Gaussian Mixture Model (HMM-GMM) architectures for speech recognition \cite{Povey2011TheKS} , complemented by separate neural modules (e.g., recurrent or transformer-based networks) for punctuation prediction \cite{Tilk2016BidirectionalRN, Courtland2020EfficientAP}. However, empirical advances in deep learning consistently demonstrate that end-to-end neural architectures outperform component-based pipelines in overall accuracy and generalizability.

The introduction of Whisper \cite{radford2023robust} , a unified neural model combining acoustic feature extraction with autoregressive language modeling, exemplifies this paradigm shift. By jointly optimizing acoustic and linguistic representations, Whisper directly generates grammatically coherent, punctuated text from raw audio signals, obviating the need for cascaded subsystems. Despite its advancements, Whisper exhibits limitations inherent to autoregressive sequence-to-sequence models:

\begin{enumerate}
\item \textbf{Hallucination artifacts:} The model occasionally produces semantically inconsistent or contextually implausible outputs, despite syntactic correctness.
\item \textbf{Computational inefficiency:} Autoregressive token-by-token decoding imposes significant latency, hindering real-time applications.
\end{enumerate}

To mitigate these constraints, subsequent work proposed WhisperX \cite{bain23_interspeech}, a refined framework incorporating algorithmic optimizations such as non-autoregressive parallel decoding and constrained beam search. These innovations aim to enhance both transcription accuracy (reducing hallucination rates) and inference speed, addressing critical bottlenecks in production-scale deployment.

\subsection{Overview of WhisperX}

WhisperX employs a multi-step architecture for ASR, beginning with Voice Activity Detection (VAD) using the pyannote.audio model \cite{Bredin23}. This model utilizes parameters such as onset and offset thresholds, as well as durations for speech detection, to effectively pinpoint the presence of speech in an audio stream. The VAD process entails several stages, including prediction of speech probability, binarization into speech and non-speech segments, and smoothing to eliminate noise and short pauses.

Following VAD, WhisperX adopts a Cut \& Merge Strategy for audio preprocessing. This method segments long speech parts into optimal chunks, allowing for parallel processing without exceeding 30 seconds in duration on segments of minimal speech activity. Thus, WhisperX enhances efficiency while minimizing errors at segment boundaries.

\subsection{Key Differences with Our Proposed Architecture}

While WhisperX features innovative strategies for maintaining accurate transcription and efficient parallel processing, our proposed architecture introduces two crucial differences that substantially enhance its performance in reducing errors and hallucinations.

\subsubsection{VAD Implementation through Wav2Vec2}

Our solution implements Voice Activity Detection (VAD) through the Wav2Vec2 model, which provides a more nuanced analysis of audio signals and a better understanding of acoustics compared to the fixed threshold approach used in WhisperX.

\subsubsection{Additional Filtering Using Audio Spectrogram Transformer (AST)}

Unlike WhisperX, which applies VAD only prior to transcription, our architecture incorporates a filtering step after the initial recognition phase using the Audio Spectrogram Transformer (AST). This enhances the validity of the segments sent to Whisper for final transcription, significantly reducing the likelihood of hallucinations.

\subsubsection{Consistency Check Between Whisper and Wav2Vec2 Outputs}

Additionally, we compare the transcription results from the Whisper model with the initial output from Wav2Vec2 to mitigate potential inaccuracies. This verification step, absent in WhisperX, serves as a potent mechanism to further minimize errors, ensuring that the system produces reliable and contextually appropriate transcriptions.

\section{Uncertainty modeling}

An uncertainty in transcription (word-wise or segment-wise) may be beneficial in some use cases:
\begin{enumerate}
    \item \textbf{Highlighting} uncertain places allows for a quick manual correction without the need to read the whole transcription.
    \item \textbf{Refusing} to transcribe some hard to hear phrases based on uncertainty scores is a useful strategy. Incorrect transcriptions can disrupt subsequent LLM-based text summarization and potentially harm an individual's reputation.
    \item \textbf{Correcting} transcriptions using subsequent stages such as language models may be more effective if we provide uncertainty scores or different transcription options.
\end{enumerate}

Uncertainty modeling is a vast area of research. In a current work we compare only the most straightforward methods that we describe in details later:
\begin{enumerate}
    \item Token scores (output probabilities) from Whisper.
    \item Disagreement between the predictions of the two pipeline stages: Whisper and Wav2Vec2. While we use Wav2Vec2 primarily for segmenting a long audio, we can make use of its predictions in uncertainty modeling.
    \item Disagreement between the Whisper predictions, obtained from the original and stretched audio. For now we preferred audio stretching over other Test-Time Augmentation (TTA) methods, as well as Monte Carlo Dropout. Their comparison may be a future work.
\end{enumerate}

\subsection{Computational efficiency}

At first glance it seems that the first option is the most computationally efficient. However, the Wav2Vec2 stage may increase the efficiency of the whole pipeline: it helps to split audio pretty quickly, and further Whisper can be run in parallel on all segments, in contrast to the Whisper long-form transcription that is sequential. After applying Wav2Vec2, we obtain its predictions for free. The third method, while requires multiple Whisper runs, is not so costly if the GPU is not fully loaded, since we can perform TTA in parallel using batching.

\subsection{Model disagreement}

Let we have transcriptions from the base (usually better) and additional (usually worse) model, e.g. from Whisper and a lightweight Wav2Vec2 segmenter. We perform the following stages:
\begin{enumerate}
    \item \textbf{Aligning} a pair of transcriptions with sequence matching, and find all differences (insertions, deletions and replacements).
    \item \textbf{Splitting or merging} the differences to achieve better linguistic matching. For example, a sequence matcher identifying the replacement "Hello Richie" -> "Richard" is split into the deletion of "Hello" and the replacement "Richie" -> "Richard." Conversely, if it finds the deletion of "no" followed by "thing" -> "nothing," we merge these into "no thing" -> "nothing."
    \item \textbf{Optional stage: applying some heuristics.} For example, we drop a replacement X -> Y if X consists only of English letters, and Y consists only of Russian letters, since it is probably a transliteration, where both options are valid. Dropping means that we accept the variant from the base model. This helps to reduce the number of differences that is usually too large.
    \item \textbf{Optional stage: LM validation.} To reduce errors from additional models, we focus on cases where the language model aligns with the additional model, i.e., the variant from the base model provides better sequence score. This approach reduces the amount of differences. Additionally, we employ a look-ahead algorithm to account for dependent subsequent differences.
\end{enumerate}

\subsection{Whisper scores}

Whisper provides probabilities for each output token. While it has been noted that models are usually overconfident in their predictions, even if they are wrong \cite{Lakshminarayanan2017}, this problem is alleviated in robust models \cite{Grabinski2022Dec}. We aim to estimate the effectiveness of Whisper probabilities as an uncertainty measure.

Whisper tokens are byte sequences of utf-8 encoding, and some utf-8 symbols can be split between two tokens. We designed an algorithm that finds Whisper token indices corresponding to each word. For example, the Russian word “\foreignlanguage{russian}{ сети}”, starting with a space, consists of two tokens (“ \foreignlanguage{russian}{с}”, “\foreignlanguage{russian}{ети}”), along with their log-probabilities. Since we use word-based uncertainty, we need to reduce these probabilities using \textit{min}, \textit{sum} or \textit{mean} operation, and empirically \textit{min} and \textit{sum} perform on par, and better than \textit{mean}.

It is worth noting that sum of log-probabilities is mathematically a log-probabilities of the whole word, up to a certain tokenization. For example, “ cat”, “ Cat”, “Cat” and “ C”+“at” are different token sequences in Whisper, and the probability of the spoken word “cat” is distributed between them. We didn't take this into account, leaving it for a future work.

After obtaining a score for each word, we select some threshold to mark each word as either certain or uncertain. Comparing to the model disagreement, here we do not have another suggestions for uncertain words (however, we could in principle extract them from Whisper).

\section{Experiments}

\begin{table*}[]
  \centering
  \begin{tabular}{|l|ll|ll|}
    \hline
    \multirow{2}{*}{\textbf{Model}} & \multicolumn{2}{l|}{\textbf{Quiet noises}}                               & \multicolumn{2}{l|}{\textbf{Loud noises (SNR = 1 dB)}}                   \\ \cline{2-5} 
    & \multicolumn{1}{l|}{WER ↓}           & {BERT-F1 ↑}       & \multicolumn{1}{l|}{WER ↓}           & {BERT-F1 ↑}       \\ \hline
    Whisper-Large-V3                & \multicolumn{1}{l|}{\textbf{0.0931}} & \textbf{0.9661} & \multicolumn{1}{l|}{0.2409}          & 0.9151          \\ \hline
    Whisper-Podlodka-V3             & \multicolumn{1}{l|}{0.1199}          & 0.9644          & \multicolumn{1}{l|}{\textbf{0.2119}} & \textbf{0.9169} \\ \hline
  \end{tabular}
  \caption{Whisper-Large-V3 and Whisper-Podlodka-V3 comparison in best ASR pipeline}
  \label{table:pisets}
\end{table*}

\begin{table*}[]
\centering
\begin{tabular}{|l|l|l|}
\hline
\textbf{Metrics} & \textbf{Pisets} & \textbf{WhisperX} \\ \hline
WER ↓            & \textbf{0.1065}           & 0.1683             \\ \hline
BERT-score ↑     & \textbf{0.9652}  & {0.9479}    \\ \hline
\end{tabular}
\caption{WhisperX and Pisets testing results on long audio lectures dataset}
\label{table:whisperx}
\end{table*}

\subsection{Lexical and semantic quality of speech recognition}

Evaluating speech recognition systems' quality is crucial due to their diverse applications, from voice assistants to transcription services. While traditional measures like Word Error Rate (WER) have been common, they may not adequately assess modern autoregressive generative decoders.

The main limitation of WER is that these systems can produce semantically accurate output that differs lexically from the original speech, which is vital in sensitive contexts like medical or legal documentation. Therefore, semantic quality measures such as BERT score (F1) are recommended, as they measure the semantic similarity between generated text and the original.

Additionally, real-world recordings often encounter noise, which can adversely affect recognition quality. Experimental evaluations should simulate various noise levels and types to better understand system performance across different acoustic environments.

In summary, a comprehensive assessment of speech recognition systems should incorporate both lexical measures like WER and semantic measures such as BERT score (F1) for a more complete understanding of their effectiveness.

\subsection{Experimental evaluation of ASR quality}

We experiment on seven long 20-40 minute Russian audios collected as a test set for our ASR system. The audios belong to different lexical and speech domains; they are parts of several Russian scientific lectures on various subjects: philology, mathematics, history, etc.

All recordings were made in relatively quiet acoustic environments typical of lecture halls; however, some background noises, such as the sound of chalk hitting a blackboard, were present. To simulate more noisy conditions, we mixed the recordings with speech-like and musical noise at a signal-to-noise ratio of 1 dB.

Table \ref{table:pisets} presents comparative results from various configurations of the Whisper architecture within the Pisets system, while table \ref{table:whisperx} details the comparative performance outcomes between the Pisets and WhisperX architectures. Based on these results, it can be inferred that the Pisets architecture provides higher recognition quality compared to WhisperX. Notably, the Whisper-Podlodka model within the Pisets architecture slightly falls short of the original Whisper-Large model under favorable acoustic conditions but begins to demonstrate advantages as the levels of background speech-like and musical noise increase.

\begin{figure*}[ht]
  \centering
  \includegraphics[width=\linewidth]{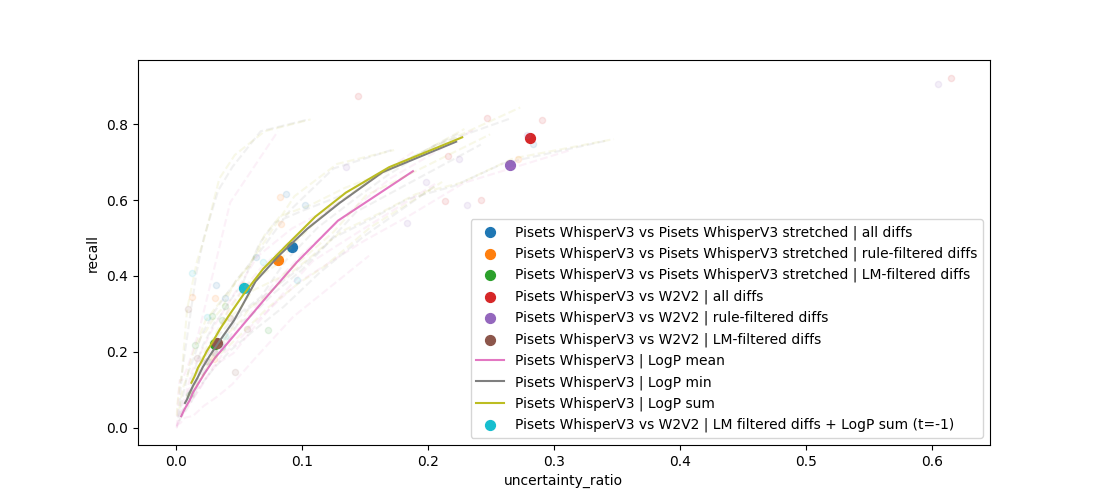} 
  \caption{The error detection recall and uncertainty ratio of different uncertainty estimation methods. The results are averaged across 7 long Russian audios, and the results for individual audios are shown in semi-transparent. Whisper scores method is show as a line for different score threshold. All model disagreement and ensembling methods cannot reliably outperform Whisper scores as a source of uncertainty. It can be seen that if we mark only around 5\% words as uncertain, we can accumulate in them 35\% of all errors (excluding errors caused by missed words in transcription).}
  \label{fig:uncertainty}
\end{figure*}

\subsection{Uncertainty modeling metrics}

It is common to evaluate uncertainty via error-retention curves \cite{Lakshminarayanan2017}, when we drop a variable percent of least-certain predictions and evaluate a quality on others, using some metric of interest. However, in long-form speech recognition, it is not clear how to evaluate WER when ignoring some words. We therefore rely on another metrics.

Let we have a list of predicted words and a boolean flag for each word (certain or uncertain) \footnote{Instead of boolean flags we could use scores and evaluate something like ROC AUC, but some methods (such as model disagreement) do not provide scores.}. We align them to ground truth words, we find incorrectly predicted words, i.e. words that correspond to “delete” or “replace” operations. We thus form a target for each word: is it correct or not? In this way, the problem is reduced to binary classification. We select two metrics that allow us to construct a Pareto-optimal frontier:
\begin{enumerate}
    \item \textbf{Uncertainty ratio}: the ratio of all predicted words marked as uncertain.
    \item \textbf{Recall of error detection}: the ratio of all incorrect words marked as uncertain.
\end{enumerate}

Note that all these calculations do not take into account the ground truth words that are not predicted by the model, since we cannot mark as uncertain a word that is not predicted. In theory, this allows the model to cheat our uncertainty metrics by predicting only a small number of the most confident words, along with the definitely incorrect words. However, this will hurt WER that is the main metric of interest.

\subsection{Uncertainty modeling experiments}

This experiments section consisted of the following pipeline:
\begin{enumerate}
    \item \textbf{Our Wav2Vec2 model} as segmenter and the additional source of predictions;
    \item \textbf{Whisper-Large-v3} as the base source of predictions and token scores;
    \item \textbf{Whisper-Large-v3} accepting strecthed words as the additional source of predictions. We use a simple audio resampling using polyphase filtering with upsampling by the factor 3 and downsampling by the factor 4. Thus, the audio is stretched by 33\%, and the pitch of the voice also changes.
\end{enumerate}

We also tried to ensemble the uncertainty mask from Whisper scores and model disagreement, considering the word as uncertain if at least one mask marks it so.

Fig. \ref{fig:uncertainty} shows the average results. No model disagreement methods consistently outperform Whisper scores as a source of uncertainty due to the limited test set size. However, marking only about 5\% of words as uncertain can capture 35\% of all errors (excluding those from missed words), making this approach very practical.

For now we use the uncertainty only for highlighting dubious places in the transcription (see Appendix \ref{appendix:uncertainty}). We also conducted preliminary experiments on feeding the text in into LLM, supplemented with instructions to resolve the disagreements based on linguistic knowledge and common sense. The experiments have shown that this may reduce WER, however is beyond the scope of the current work.

\section{Conclusion}

This paper presents a novel framework aimed at improving speech recognition systems, addressing challenges such as hallucinations, domain adaptability, and acoustic-linguistic variability. The combination of Wav2Vec2 for speech segmentation, AST for false positive filtering, and Whisper for final transcription significantly reduced errors across various acoustic conditions. The integration of diverse Russian speech corpora, along with the use of the BIRM model for fine-tuning, further enhanced the system's robustness to unfamiliar domains.

Additionally, the implementation of advanced uncertainty modeling techniques provided practical recommendations for improving transcription quality. These enhancements led to the development of a reliable system capable of delivering high-quality transcription in a variety of scenarios, including automatic dictation and conversational AI systems.

Future work is planned to expand uncertainty handling capabilities and enhance adaptation to multilingual datasets, allowing for more effective recognition of English speech by non-native speakers, as well as the recognition of Bengali, Spanish, and other languages.

\section{Limitations}

Our system currently demonstrates insufficient performance when addressing the recognition of homophones and words or phrases that exhibit similar phonetic characteristics. To enhance the efficacy of speech recognition in such scenarios, it is imperative to incorporate not only semantic but also pragmatic levels of understanding within the system. In the context of generative autoregressive models, the pragmatic level can be delineated through instructions (prompts) that elucidate the local conversational context and specify the key terminology employed by the interlocutors. Unfortunately, architectures akin to Whisper exhibit limitations in their capacity to adhere to these instructions. Consequently, to address the challenge of effectively integrating pragmatics into the speech recognition system, we plan to incorporate large multimodal models, such as Qwen-Audio.

\section{Acknowledgements}

The work is supported by the grant for the implementation of the strategic academic leadership program "Priority 2030" at Novosibirsk State University.

\bibliography{custom}

\appendix

\section{Dictation mistakes overview}

On April 20, 2024, our ASR system participated in the “Total Dictation” \cite{totaldict} event along with other writers. “Total Dictation” is an annual mass event in Russia where thousands of participants write down a text read by a narrator. 

\subsection{Acoustic conditions}

The dictation took place in a 200-person classroom with a microphone and the text was read by a professional philologist. The narrator pronounced the text clearly and loudly, which was favorable for the recognition process. The room where the dictation took place had background noise due to the presence of over a hundred participants. Conversations, noise from people moving, coughing, and rustling paper all created acoustic noise that hindered speech recognition. The large auditorium where the dictation was held had high reverberation, which negatively affected the audibility of speech. The input signal was obtained by classroom microphone, which recorded speech according the acoustics of the room.

\subsection{Linguistic Conditions of the Text}

The text was written in Russian in a free, conversational style. It was dedicated to the topic of diaries and their role in a person's life. The text's lexicon was straightforward, using common words and expressions. The text had a clear structure, consisting of several paragraphs. 

First of all, the text was read entirely, then each sentence was repeated at a fast pace. After that it was dictated slowly by parts, sometimes the parts were repeated at the request of the listeners. After all, the sentence was repeated in full at a fast pace. The narrator inserted additional comments into the text that did not require transcription. This added the task of separating the main text from extraneous comments. Each paragraph was announced with phrases like “We start the next sentence with a new line” or “Let's start a new paragraph”. At the end of the dictation, the text was repeated once more at a fast pace. The narrator also made some comments not related to the content of the text. For example, “Let's take a break and warm our fingers, like we did in school” or “Be patient, the end is near”.

To detect insertions we have trained the Longformer model~\cite{beltagy2020longformer}. As a dataset, out-of-context inserts and line break inserts were generated in texts. The text recognised at the first dictation reading with all the inserts in the post-processing was run through the Longformer. It was not possible to remove a sufficient number of inserts, but it split the text into paragraphs correctly. Then the text was recognized, which was repeated by the speaker in the second reading without inserts. The line break flags were taken from the first text with inserts and applied to the second text without inserts. Thus, the text without inserts and with line breaks in the right places was obtained.

\subsection{Typology of model mistakes}

Based on the results of the dictation, the following observations about the model work were made:

\begin{enumerate}
    \item Two spelling errors were made. Both related to the endings of a noun (\foreignlanguage{russian} {“портрет гимназистке”} — genitive singular) and an adjective (\foreignlanguage{russian} {“ярко-синями”}) and also three punctuation errors (direct speech, homogeneous parts of a sentence, comparative turnover).
    \item Eight words (total count 276) (\foreignlanguage{russian} {“рук”, “маскарады”, “разумеется”, “в мире почерком”, “модным”, “приходило”}) were missed at the end of sentences. In this case, model did not put a full stop, starting the next sentence with a capital letter. Most of the omissions lead to a violation of the sentence structure.
    \item The ASR system ignored the parceling that occurred twice in the text, although the narrator drew attention to it. For example, the last sentences of the text were combined into one: \foreignlanguage{russian} {“Главное, чего не следовало делать, это вырывать исписанные страницы. Отказываться от своего прошлого”}. However, in both cases, punctuation marks were placed correctly, and such a case would not have been counted as an error when checking other writers.
    \item In eight cases, the ASR system made “mishearings”, writing down words that sounded close but in most cases were far in meaning from the original ones: instead of \foreignlanguage{russian} {“клеенчатых” — “кальиончатых”, “чернилами” — “черепами”, “катки” — “ходки”, “хранились”  — “хоронились”, “наивысшего” — “наявившего”, “свадьбой” — “спать”}. It should be noted that the words \foreignlanguage{russian} {“клеенчатых”} and \foreignlanguage{russian} {“почерком”} caused the greatest difficulties for other dictation writers. The construction \foreignlanguage{russian} {“читай – не хочу”}, which the model recorded as \foreignlanguage{russian} {“Считай, не хотите”}, was not recognized by the model.
    \item We will separately point out the “mishearing” that led to the fact that the content of the sentence was violated, but a similar error is common among others who wrote the text: instead of \foreignlanguage{russian} {“Она мечтала о славе и так смело открывалась в своих записях…”} it was \foreignlanguage{russian} {“Она мечтала о славе, и та смело открывалась в своих записях…”}.
\end{enumerate}

Overall, the “model” copes well with spelling and punctuation rules, ignores repetitions of parts of sentences and words not related to the content of the text, and correctly places paragraphs. The number of spelling and punctuation errors made by the system is less than that of most who wrote the same text. The model is able to transform the original text without violating the rules of the Russian language. However, in some cases, the model incorrectly perceives words and expressions, mainly at the end of a sentence, omitting them or replacing them, including with non-existent forms. The experts of “Total Dictation” (professional philologists and linguists) evaluated the work of our ASR system as B (“good”). For comparison, many people write “Total Dictation” with a grade of F, making a small number of mistakes.

\section{Noisy audio testing}

The tables \ref{table:quiet} and \ref{table:noisy} show different results of ASR pipeline configurations on noisy and clean audio.

\section{Testing computational efficiency}
The table \ref{table:time} shows that using Wav2Vec2 "smart" chunking outperforms the uniform chunking of the original Whisper model in terms of inference time.

\makeatletter
\setlength{\@dblfptop}{0pt}  
\setlength{\@dblfpsep}{30pt} 
\setlength{\@dblfpbot}{0pt}  
\makeatother

\begin{table*}[]
  \centering
  \begin{tabular}{|l|l|l|}
    \hline
    \textbf{Configuration} & \textbf{WER ↓}                    & \textbf{BERT-F1 ↑}                \\ \hline
    Whisper with uniform chunking                  & 0.1995          & 0.9102          \\ \hline
    Whisper with Wav2Vec2 "smart" chunking         & \textbf{0.1065} & \textbf{0.9652} \\ \hline
    Whisper with Wav2Vec2 "smart" chunking and AST & 0.1109          & 0.9588          \\ \hline
  \end{tabular}
  \caption{Different ASR pipeline configurations' results for quiet noises audio}
  \label{table:quiet}
\end{table*}

\begin{table*}[]
      \centering
  \begin{tabular}{|l|l|l|}
    \hline
    \textbf{Configuration} & \textbf{WER ↓}                    & \textbf{BERT-F1 ↑}                \\ \hline
    Whisper with uniform chunking                  & 0.3825          & 0.8508          \\ \hline
    Whisper with Wav2Vec2 "smart" chunking         & \textbf{0.2119} & \textbf{0.9169} \\ \hline
    Whisper with Wav2Vec2 "smart" chunking and AST & 0.2133          & 0.9160          \\ \hline
  \end{tabular}
  \caption{Different ASR pipeline configurations' results for loud noises audio}
  \label{table:noisy}
\end{table*}

\begin{table*}[]
      \centering
  \begin{tabular}{|l|l|l|l|}
    \hline
    \textbf{Configuration} & \textbf{Max ↓} & \textbf{Average ↓}  & \textbf{Median ↓} \\ \hline
    Whisper with uniform chunking                  &  192.045  &  136.377 & \textbf{121.090} \\ \hline
    Whisper with Wav2Vec2 "smart" chunking         &  152.524  &  133.219 & 134.918 \\ \hline
    Whisper with Wav2Vec2 "smart" chunking and AST &  \textbf{151.923}  &  \textbf{131.495} & 130.809 \\ \hline
  \end{tabular}
  \caption{Different ASR pipeline configurations' time (in seconds) results for noised audio}
  \label{table:time}
\end{table*}

\begin{figure*}[]
  \centering
  \includegraphics[width=\linewidth]{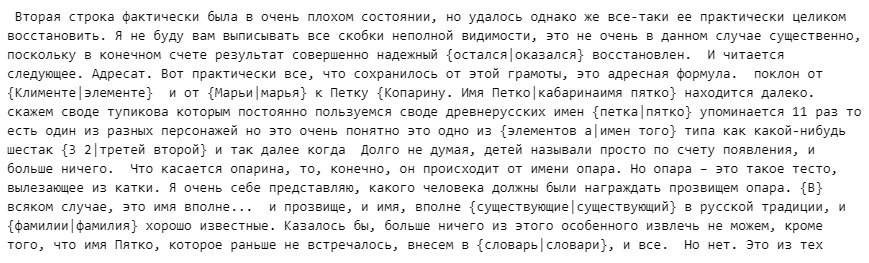}
  \caption{The example of highlighting dubious places in the transcription, based on uncertainty estimation with model disagreement.}
  \label{fig:generation_example}
\end{figure*}

\section{Uncertainty places in final transcription}
\label{appendix:uncertainty}
The example of highlighting dubious places in the transcription, based on uncertainty estimation with model disagreement are shown on Fig. \ref{fig:generation_example}.

\end{document}